\documentclass{ecai} 

\usepackage{latexsym}
\usepackage{amssymb}
\usepackage{amsmath}
\usepackage{amsthm}
\usepackage{booktabs}
\usepackage{enumitem}
\usepackage{graphicx}
\usepackage{color}
\usepackage{multirow}

\newcommand{\BibTeX}{B\kern-.05em{\sc i\kern-.025em b}\kern-.08em\TeX}

\begin{document}


\begin{frontmatter}


\paperid{48} 


\title{\textsc{KGPrune}: a Web Application to Extract Subgraphs of Interest from Wikidata with Analogical Pruning}


\author[A]{\fnms{Pierre}~\snm{Monnin}\orcid{0000-0002-2017-8426}\thanks{Corresponding Author. Email: pierre.monnin@inria.fr.}}
\author[B]{\fnms{Cherif-Hassan}~\snm{Nousradine}}
\author[B,C]{\fnms{Lucas}~\snm{Jarnac}\orcid{0000-0002-2819-2679}} 
\author[D]{\fnms{Laurel}~\snm{Zuckerman}\orcid{0000-0001-8630-0554}} 
\author[B,E]{\fnms{Miguel}~\snm{Couceiro}\orcid{0000-0003-2316-7623}}

\address[A]{Université Côte d’Azur, Inria, CNRS, I3S, Sophia-Antipolis, France}
\address[B]{Université de Lorraine, CNRS, LORIA, Nancy, France}
\address[C]{Orange, France}
\address[D]{Independent Researcher}
\address[E]{INESC-ID, Instituto Superior Técnico, Universidade de Lisboa, Lisboa, Portugal}


\begin{abstract}
Knowledge graphs (KGs) have become ubiquitous publicly available knowledge sources, and are nowadays covering an ever increasing array of domains. 
However, not all knowledge represented is useful or pertaining  when considering a new application or specific task. 
Also, due to their increasing size, handling large KGs in their entirety entails scalability issues. 
These two aspects asks for efficient methods to extract subgraphs of interest from existing KGs. 
To this aim, we introduce \textsc{KGPrune}, a Web Application that, given seed entities of interest and properties to traverse, extracts their neighboring subgraphs from Wikidata. 
To avoid topical drift, \textsc{KGPrune} relies on a frugal pruning algorithm based on analogical reasoning to only keep relevant neighbors while pruning irrelevant ones.  
The interest of \textsc{KGPrune} is illustrated by two concrete applications, namely,  bootstrapping  an enterprise KG and extracting  knowledge related to looted artworks. 
\end{abstract}

\end{frontmatter}


\section{Introduction}

Knowledge graphs (KGs) are structured representations that model the knowledge of one or several domains.
Their atomic units are triples $(h, p, o)$ that represent the existence of a relationship $p$ between two entities $h$ and $o$. 
With their flexibility and the knowledge they provide, KGs have become major assets that fuel various methods of artificial intelligence (\textit{e.g.}, retrieval augmented generation for large language models~\citep{panLWCWW24}, machine learning in general~\citep{dAmatoMMS23}) with applications in a wide array of domains (\textit{e.g.}, search, e-commerce, social networks, life sciences~\citep{chen0HJLMP0T23,hoganBCAMGKGNNNNPRRSSSZ21,noyGJNPT19}). 

In the seminal spirit of the Semantic Web~\citep{berners2001}, existing KGs are often re-used for building new KGs or for supporting new tasks or applications~\cite{dongGHHLMSSZ14,fernandez1997,mahdisoltaniBS15,swartoutPKR1996}.
This is possible due to the increasing number and size of publicly available KGs\footnote{https://lod-cloud.net/}, and in particular of large KGs covering several domains such as Wikidata. 
The latter is a generic KG of more than 100 million nodes\footnote{\url{https://www.wikidata.org/wiki/Wikidata:Statistics}} that supports Wikipedia~\citep{vrandecicK14}. 
Wikidata is considered as a premium source of knowledge but several issues hinder its reusage~\cite{jarnacM22,shbitaGLDR23}.
First, its large size entails scalability issues when handling the graph (\textit{e.g.}, storage, query performance).
Second, not all represented knowledge is relevant to the considered tasks or applications. 
For example, one of the neighboring entities of \texttt{Microsoft SharePoint} is \texttt{Dating App} which may not be of interest when building a enterprise KG modeling the IT domain. 

To address such issues, several authors have proposed extracting subgraphs, either manually with some early examples dating back to 1996~\cite{swartoutPKR1996} or automatically~\cite{jarnacCM23,jarnacM22,shbitaGLDR23}.
In particular, we recently proposed an approach that traverses the neighborhood of seed entities provided by users, keeping relevant neighbors while pruning irrelevant ones~\citep{jarnacCM23}. 
This approach relies on analogical inference and exhibits high performance, including in transfer settings, with a drastically low number of parameters. 

Building on this previous work, we propose \textsc{KGPrune},\footnote{https://kgprune.loria.fr} a Web Application to extract subgraphs from Wikidata given seed entities and properties of interest to the user. 
\textsc{KGPrune} can be used both from a browser and programmatically through an API, allowing users with various technical expertise to interact with our pruning approach. 
In the following, after describing the features and technical architecture of \textsc{KGPrune}, we illustrate its interest with two concrete applications: the bootstrapping of an enterprise knowledge graph, and the extraction of knowledge related to looted artworks. 
A video of our demonstration is available on YouTube.\footnote{https://youtu.be/mt5gF4ZmhGY}

\section{\textsc{KGPrune}: Extracting Subgraphs of Interest}

\begin{figure*}
    \centering
    \includegraphics[width=0.85\textwidth]{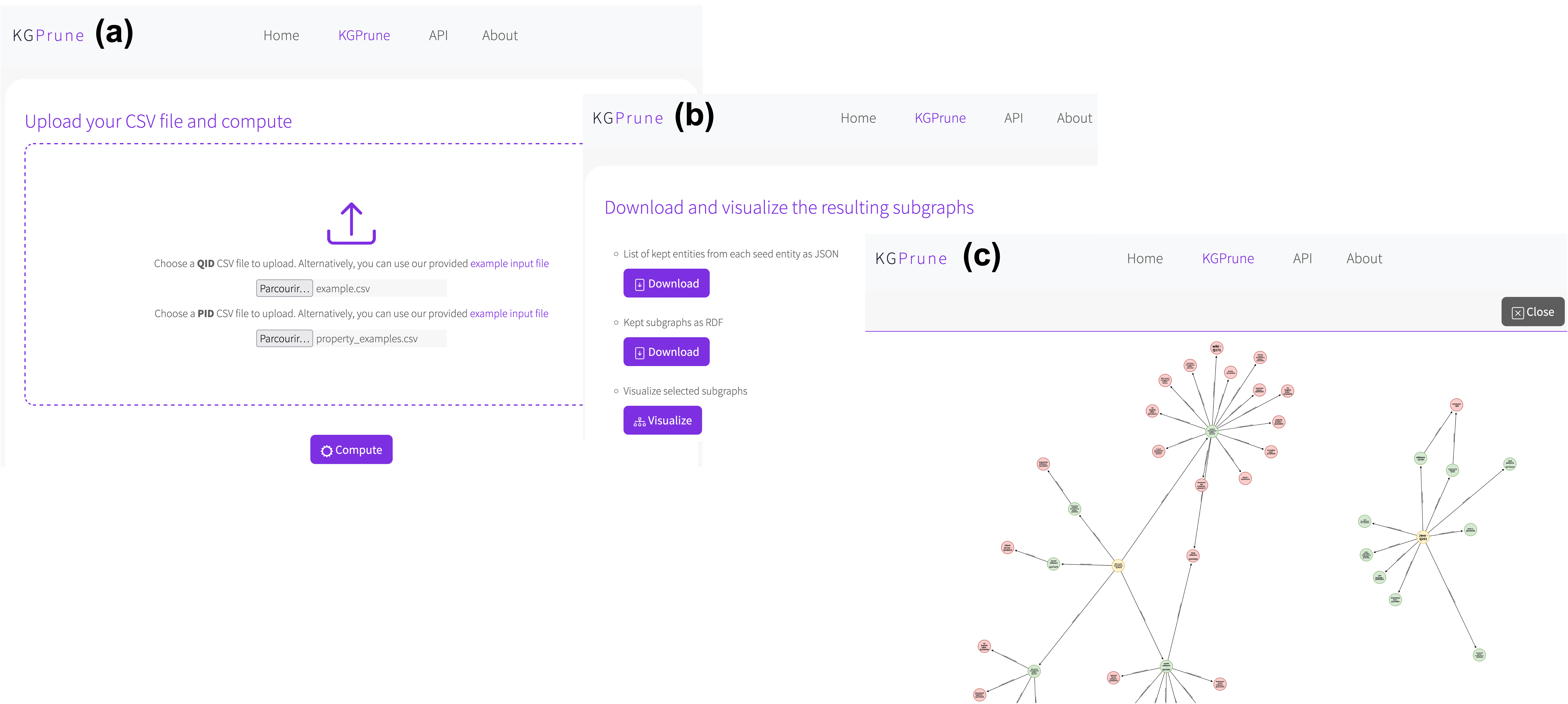}
    \caption{Main screens of \textsc{KGPrune}. (a) Upload form where two CSV files are required from the user: one indicating QIDs of seed entities and one indicating PIDs of properties to traverse. (b) Result page where the user can choose to visualize the extracted subgraphs or download them in JSON or RDF. (c) Visualization of the extracted subgraphs where seed entities are in yellow, kept neighbors in green, and pruned neighbors in red.}
    \label{fig:kgprune-webapplication}
\end{figure*}

The main screens of the \textsc{KGPrune} Web Application are presented in Figure~\ref{fig:kgprune-webapplication}.
We describe below the main characteristics and steps for interacting with the application.

\paragraph{Supporting KG.} We chose to build \textsc{KGPrune} upon the Wikidata KG as it is large and generic, and thus can serve as a premium source of knowledge for several domains. 
However, it should be noted that our approach could be applied on any KG.

\paragraph{Input files.} \textsc{KGPrune} only requires as input from the user two CSV files, as illustrated in Table~\ref{tab:input-files}. 
The file \texttt{qid\_example.csv} contains QIDs identifying seed entities of interest whose neighborhood will be retrieved. 
Here, as an example, we consider \texttt{Microsoft SharePoint} (Q18833) and the \texttt{Java} programing language (Q251).
The file \texttt{pid\_example.csv} contains PIDs identifying properties of interest whose edges will be traversed. 
Here, we consider \texttt{instance of} (P31), \texttt{subclass of} (P279), and \texttt{part of} (P361). 
Note that indicating the PID of a property leads to traversing direct edges whereas indicating (-)PID leads to traversing inverse edges. 
Here, both direct and inverse P279 edges will be traversed. 
The upload screen of \textsc{KGPrune} is presented in Figure~\ref{fig:kgprune-webapplication}a. 

\begin{table}
\caption{Example of input files for \textsc{KGPrune}. The file \texttt{qid\_example.csv} contains QIDs of seed entities of interest. Their neighborhood will be retrieved by traversing edges labeled by the properties whose PIDs are specified in \texttt{pid\_example.csv}. }
\label{tab:input-files}
\centering
\begin{tabular}{l|l} 
\toprule
\texttt{qid\_example.csv} & \texttt{pid\_example.csv} \\
\midrule
Q18833 & P31  \\
Q251 & P279 \\
& (-)P279 \\
& P361 \\
\bottomrule
\end{tabular}
\end{table}

\paragraph{Subgraph extraction.} After input CSV files have been uploaded, \textsc{KGPrune} executes our traversal and pruning algorithm~\citep{jarnacCM23}\footnote{\url{https://github.com/Orange-OpenSource/analogical-pruning}}.
Starting from seed entities, edges labeled by the specified properties are traversed. 
For each neighbor, our analogical pruning model decides either to keep or prune it.
Analogies are statements of the form ``A is to B as C is to D'', modeled as quadruples $A:B::C:D$ such as $\text{Paris}:\text{France}::\text{Berlin}:\text{Germany}$. 
Such quadruples capture similarities and dissimilarities between objects~\cite{miclet2008,mitchell21}.  
Here, given a seed entity $e_s^u$ specified by the user and one of its neighbors $e_r^u$, our model predicts whether they form an analogy with a seed entity $e_s^k$ and one of its neighbor $e_r^k$ for which a ``keep'' decision is known:
$$\underbrace{e_s^k~:~e_r^k}_{\text{Known ``keep'' decision}}~::~\underbrace{e_s^u~:~e_r^u}_{\text{Unknown decision}}$$
\noindent This prediction relies on the pre-learned embeddings of the entities and the convolutional model for analogy detection introduced by \citet{lim19}. 
With its architecture, the analogy-based model is able to capture relative similarities and dissimilarities between seed entities and their neighbors to keep or to prune, and thus is able to generalize to heterogeneous unseen entities.
If our model predicts that they form an analogy, the known decision between $e_s^k$ and $e_r^k$ (\textit{i.e.}, keep $e_r^k$) is extrapolated to $e_r^u$. 
Otherwise, $e_r^u$ is pruned.
Note that the known decisions originate from one manually annotated dataset named \texttt{dataset1} that is publicly available\footnote{\url{https://doi.org/10.5281/zenodo.8091584}\label{fn:zenodo-datasets}}.

This process is performed iteratively on the neighborhood of kept neighbors until no more neighbors can be reached. 
Results are then displayed to the user (Figure~\ref{fig:kgprune-webapplication}b) who can choose to visualize the extracted subgraphs (Figure~\ref{fig:kgprune-webapplication}c) or download them as JSON or RDF to be imported into a new KG.
The visualization interface allows users to explore the neighborhoods of the seed entities, and assess the pruning results.
In particular, users can notice in the UI if our model wrongfully pruned a neighbor of interest to the users, and add it to the seed entities to force its consideration.
This lays the path towards an iterative pruning process in which users explore pruning results and provide feedback that is leveraged in the subsequent iterations.

\begin{figure}[h!]
    \centering
\includegraphics[width=0.4\textwidth]{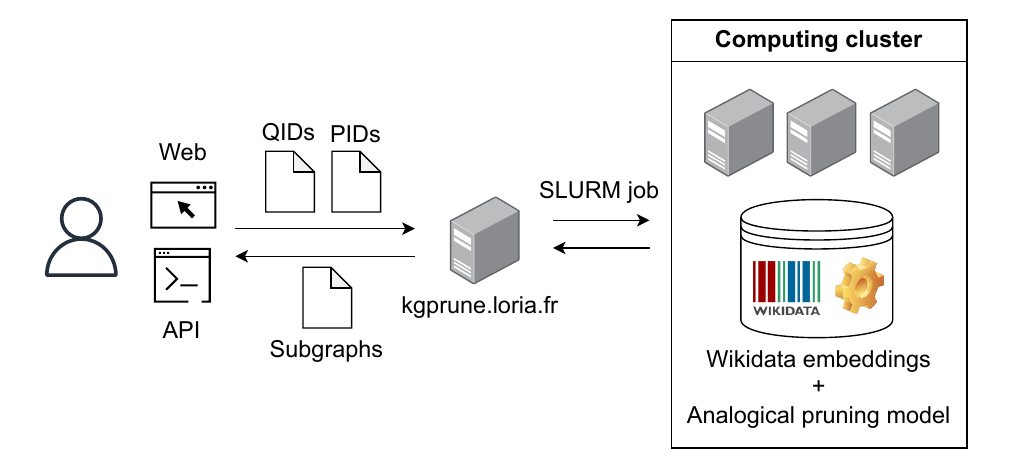}
    \caption{Technical architecture of \textsc{KGPrune}. Users can upload subgraph extraction tasks via the website or the API. These tasks are then submitted as SLURM jobs to our computing cluster.}
    \label{fig:kgprune-tech-architecture}
\end{figure}

\paragraph{Technical architecture. } \textsc{KGPrune} relies on the technical architecture presented in Figure~\ref{fig:kgprune-tech-architecture}. 
Users can interact with the application via a Web browser or the provided API. 
Their subgraph extraction tasks are sent as SLURM jobs to our computing clusters where Wikidata adjacency and pre-trained embeddings, as well as our analogical pruning models are loaded and used in inference.

For learning Wikidata embeddings, we used the TransE~\cite{bordesUGWY13} model with a dimension of 200.
For the analogical model, we trained it using using \texttt{dataset1} among two manually annotated datasets publicly available\footref{fn:zenodo-datasets}. 
We use 16 filters on the first convolutional layer and 8 filters on the second convolutional layer. 

Our model achieves competitive performance compared to the main models of the state of the art, with a drastically lower number of parameters and a superior generalization capability in a transfer setting (Table~\ref{tab:results}).

\begin{table}
\caption{Performance of our model compared to LSTM, its main competitor. The transfer setting corresponds to training the model on \texttt{dataset1} and testing it on \texttt{dataset2}. Experiments on each dataset were performed using 5-fold cross-validation and hypertuning the number of filters. Full results are available in~\cite{jarnacCM23}.}
\label{tab:results}
\centering
\begin{tabular}{ll|rr} 
\toprule
\multicolumn{2}{l|}{Model} & \multicolumn{1}{c}{LSTM} & \multicolumn{1}{c}{Path Analogy} \\
\midrule
\multirow{4}{*}{\texttt{dataset1}} & Precision & $79.72 \pm 5.17$  & $80.10 \pm 0.84$ \\
& Recall & $76.00 \pm 6.59$ & $74.44 \pm 5.28$ \\
& F1 & $77.43 \pm 2.38$ & $77.06 \pm 2.89$ \\
& ACC & $83.48 \pm 3.05$ & $83.51 \pm 2.87$ \\
& \# parameters & 210,751 & 1,401 \\
\midrule
\multirow{4}{*}{\texttt{dataset2}} & Precision & $78.49 \pm 8.80$  & $81.63 \pm 8.27$ \\
& Recall & $94.58 \pm 2.96$ & $94.90 \pm 2.16$ \\
& F1 & $85.36 \pm 4.53$  & $87.54 \pm 5.05$ \\
& ACC & $78.66 \pm 5.95$ & $82.50 \pm 6.07$ \\
& \# parameters & 210,751 & 251 \\
\midrule
\multirow{4}{*}{Transfer setting} & Precision & $92.83$ & $91.49$ \\
& Recall & $74.73$ & $83.39$ \\
& F1 & $82.80$ & $87.25$\\
& ACC & $80.04$ & $84.33$ \\
\bottomrule
\end{tabular}
\end{table}

\section{Illustrative Use Cases}

To showcase the impact of \textsc{KGPrune}, we experimented on two use cases, namely, enterprise KG bootstrapping and extracting subgraphs related to looted artworks, allowing us to attest the usefulness of our tool on distinct real-world applications. 

\paragraph{Bootstrapping an Enterprise Knowledge Graph (EKG). } Building a new KG requires its bootstrapping with a high quality nucleus that can then support automatic knowledge extraction approaches from structured or unstructured data (\textit{e.g.}, tables, texts)~\cite{liuCTHLM23,sequedaL2021,weikumDRS21}. 
Indeed, these approaches then enrich the KG while being guided by the terms and relations the KG provides, forming a virtuous loop. 

To build such a nucleus, several authors rely on Wikidata.
To limit the size of the created nucleus, they select parts of the neighborhood of seed entities of interest with a distillation~\cite{shbitaGLDR23} or a pruning process~\cite{jarnacM22,jarnacCM23}. 
Their traversal of the graph focuses on the ontology hierarchy, only upward~\cite{shbitaGLDR23} or both upward and
downward~\cite{jarnacM22}. 

In~\cite{jarnacCM23}, we proposed our pruning approach to bootstrap an EKG focused on the IT domain, traversing the ontology upward and downward starting from seed entities of interest available in the company internal glossary. 
The competitive performance with low complexity obtained by our approach (Table~\ref{tab:results}) illustrates its interest for this use case.
With \textsc{KGPrune}, we extended our previous approach by allowing the user to define the properties to traverse,  enriching the subgraphs extracted from Wikidata, and with visualization capabilities to let user explore extracted neighborhoods.

\paragraph{Extracting Subgraphs Related to Looted Artworks.} 
Art looting networks operate on many hidden levels over long periods
of time.
Some agencies emphasize that it is
a criminal industry grossing in the billions annually. Reliable documentation is
of utmost importance to finding lost or stolen cultural property, and to establishing rightful ownership. This is however a challenging task \citep{finkSK14} since ``the data
on cultural heritage is locked up in data silos making it exceptionally
difficult to search, locate, and obtain reliable documentation''. 

The authors of \cite{finkSK14} propose the use of Linked Open Data (LOD) as a global database on cultural
heritage, and explored the potential of LOD to integrate  large quantities of cultural heritage data to facilitate access to information in this domain. In turn, this could help to protect cultural property from looting, as well as track looted artworks \citep{zuckerman20,zuckerman21}.
However, stolen art tracking involves knowledge pertaining to artworks, genealogy, ownership, provenance, some of which is present in generic KGs together with irrelevant knowledge to the present task (\textit{e.g.}, biology, computer science). 
For example, Wikidata contains 43,730 art dealers, collectors, curators, and galleries; 6,648 art museums; 930,405 paintings; 251 persons investigated by the Art Looting Investigation Unit\footnote{https://www.wikidata.org/wiki/Q30335959}; and properties such as \texttt{owner of} and \texttt{owned by}.
Hence, the need to extract specific subgraphs addressing relevent themes while avoiding false, inaccurate, or irrelevant information.

In this view, \textsc{KGPrune} has the potential to extract and collect trustworthy and pertinent information from Wikidata.
In preliminary experiments, we applied our approach on the neighborhood of known artworks (\textit{e.g.} \texttt{Cypresses}), artists (\textit{e.g.}, \texttt{Alexej von Jawlensky}), museums (\textit{e.g}, \texttt{National Gallery of Arts}), and art dealers (\textit{e.g.}, \texttt{Alfred Flechtheim}).
Results showed good alignment with human needs when detecting neighbors relevant to information needed for tracking stolen art. 
We are in the process of exploring further how \textsc{KGPrune}, and especially its pruning and visualization features, can support other use cases related to cultural heritage. 

\section{Conclusion and Perspectives}

In this paper, we presented \textsc{KGPrune}, a Web Application allowing users to extract subgraphs of interest from Wikidata by providing seed entities of interest and properties to traverse.
Our application prevents potential topical drift when traversing the graph by relying on an efficient analogy-based pruning mechanism. 
Users can interact with \textsc{KGPrune} via a Web browser and an API, which enables its to seamless integration in various working pipelines.
We demonstrated the interest of the application with two concrete use cases.

At present, \textsc{KGPrune} only supports Wikidata.
In the future, we envision to integrate additional KGs (\textit{e.g.}, DBpedia~\cite{lehmannIJJKMHMK15}, YAGO~\cite{tanonWS20}, Bio2RDF~\cite{dumontierCCAEBD14}) providing users an enhanced context from which extract subgraphs. 
Additionally, our analogy-based model is trained on a manually annotated dataset of seed entities and neighbors to keep or prune.
Even if experiments and use cases highlighted the generalization capability of our model, it may be possible that the definition of kept and pruned neighbors learned does not apply well to other applications.
To address this question, we plan on allowing users to provide their own examples of kept and pruned neighbors.
These examples could be used in the inference phase or even to train tailored models on-the-fly, given the reduced complexity of our models.


\begin{ack}
This work is supported by the AT2TA project (\url{https://at2ta.loria.fr/}) funded by the French National Research Agency (``Agence Nationale de la Recherche'' – ANR) under grant ANR-22-CE23-0023.
\end{ack}


\bibliography{bibliography}

\begin{thebibliography}{27}
\providecommand{\natexlab}[1]{#1}
\providecommand{\url}[1]{\texttt{#1}}
\expandafter\ifx\csname urlstyle\endcsname\relax
  \providecommand{\doi}[1]{doi: #1}\else
  \providecommand{\doi}{doi: \begingroup \urlstyle{rm}\Url}\fi

\bibitem[Berners-Lee et~al.(2001)Berners-Lee, Hendler, and Lassila]{berners2001}
T.~Berners-Lee, J.~Hendler, and O.~Lassila.
\newblock The semantic web.
\newblock \emph{Scientific american}, 284\penalty0 (5):\penalty0 28--37, 2001.

\bibitem[Bordes et~al.(2013)Bordes, Usunier, Garc{\'{\i}}a{-}Dur{\'{a}}n, Weston, and Yakhnenko]{bordesUGWY13}
A.~Bordes, N.~Usunier, A.~Garc{\'{\i}}a{-}Dur{\'{a}}n, J.~Weston, and O.~Yakhnenko.
\newblock Translating embeddings for modeling multi-relational data.
\newblock In \emph{Advances in Neural Information Processing Systems 26: 27th Annual Conference on Neural Information Processing Systems 2013. Proceedings of a meeting held December 5-8, 2013, Lake Tahoe, Nevada, United States}, pages 2787--2795, 2013.
\newblock URL \url{https://proceedings.neurips.cc/paper/2013/hash/1cecc7a77928ca8133fa24680a88d2f9-Abstract.html}.

\bibitem[Chen et~al.(2023)Chen, Dong, Hastings, Jim{\'{e}}nez{-}Ruiz, L{\'{o}}pez, Monnin, Pesquita, Skoda, and Tamma]{chen0HJLMP0T23}
J.~Chen, H.~Dong, J.~Hastings, E.~Jim{\'{e}}nez{-}Ruiz, V.~L{\'{o}}pez, P.~Monnin, C.~Pesquita, P.~Skoda, and V.~A.~M. Tamma.
\newblock Knowledge graphs for the life sciences: Recent developments, challenges and opportunities.
\newblock \emph{{Transactions on Graph Data and Knowledge}}, 1\penalty0 (1):\penalty0 5:1--5:33, 2023.
\newblock \doi{10.4230/TGDK.1.1.5}.
\newblock URL \url{https://doi.org/10.4230/TGDK.1.1.5}.

\bibitem[d'Amato et~al.(2023)d'Amato, Mahon, Monnin, and Stamou]{dAmatoMMS23}
C.~d'Amato, L.~Mahon, P.~Monnin, and G.~Stamou.
\newblock Machine learning and knowledge graphs: Existing gaps and future research challenges.
\newblock \emph{{Transactions on Graph Data and Knowledge}}, 1\penalty0 (1):\penalty0 8:1--8:35, 2023.
\newblock \doi{10.4230/TGDK.1.1.8}.
\newblock URL \url{https://doi.org/10.4230/TGDK.1.1.8}.

\bibitem[Dong et~al.(2014)Dong, Gabrilovich, Heitz, Horn, Lao, Murphy, Strohmann, Sun, and Zhang]{dongGHHLMSSZ14}
X.~Dong, E.~Gabrilovich, G.~Heitz, W.~Horn, N.~Lao, K.~Murphy, T.~Strohmann, S.~Sun, and W.~Zhang.
\newblock Knowledge vault: a web-scale approach to probabilistic knowledge fusion.
\newblock In \emph{The 20th {ACM} {SIGKDD} International Conference on Knowledge Discovery and Data Mining, {KDD} '14, New York, NY, {USA} - August 24 - 27, 2014}, pages 601--610. {ACM}, 2014.
\newblock \doi{10.1145/2623330.2623623}.
\newblock URL \url{https://doi.org/10.1145/2623330.2623623}.

\bibitem[Dumontier et~al.(2014)Dumontier, Callahan, Cruz{-}Toledo, Ansell, Emonet, Belleau, and Droit]{dumontierCCAEBD14}
M.~Dumontier, A.~Callahan, J.~Cruz{-}Toledo, P.~Ansell, V.~Emonet, F.~Belleau, and A.~Droit.
\newblock Bio2rdf release 3: {A} larger, more connected network of linked data for the life sciences.
\newblock In \emph{Proceedings of the {ISWC} 2014 Posters {\&} Demonstrations Track a track within the 13th International Semantic Web Conference, {ISWC} 2014, Riva del Garda, Italy, October 21, 2014}, volume 1272 of \emph{{CEUR} Workshop Proceedings}, pages 401--404. CEUR-WS.org, 2014.
\newblock URL \url{https://ceur-ws.org/Vol-1272/paper_121.pdf}.

\bibitem[Fernández-López et~al.(1997)Fernández-López, Gomez-Perez, and Juristo]{fernandez1997}
M.~Fernández-López, A.~Gomez-Perez, and N.~Juristo.
\newblock Methontology: from ontological art towards ontological engineering.
\newblock \emph{Engineering Workshop on Ontological Engineering (AAAI97)}, 03 1997.

\bibitem[Fink et~al.(2014)Fink, Szekely, and Knoblock]{finkSK14}
E.~E. Fink, P.~A. Szekely, and C.~A. Knoblock.
\newblock How linked open data can help in locating stolen or looted cultural property.
\newblock In \emph{Digital Heritage. Progress in Cultural Heritaage: Documentation, Preservation, and Protection - 5th International Conference, EuroMed 2014, Limassol, Cyprus, November 3-8, 2014. Proceedings}, volume 8740 of \emph{Lecture Notes in Computer Science}, pages 228--237. Springer, 2014.
\newblock \doi{10.1007/978-3-319-13695-0_22}.
\newblock URL \url{https://doi.org/10.1007/978-3-319-13695-0_22}.

\bibitem[Hogan et~al.(2021)Hogan, Blomqvist, Cochez, d'Amato, de~Melo, Gutierrez, Kirrane, Gayo, Navigli, Neumaier, Ngomo, Polleres, Rashid, Rula, Schmelzeisen, Sequeda, Staab, and Zimmermann]{hoganBCAMGKGNNNNPRRSSSZ21}
A.~Hogan, E.~Blomqvist, M.~Cochez, C.~d'Amato, G.~de~Melo, C.~Gutierrez, S.~Kirrane, J.~E.~L. Gayo, R.~Navigli, S.~Neumaier, A.~N. Ngomo, A.~Polleres, S.~M. Rashid, A.~Rula, L.~Schmelzeisen, J.~Sequeda, S.~Staab, and A.~Zimmermann.
\newblock \emph{Knowledge Graphs}.
\newblock Synthesis Lectures on Data, Semantics, and Knowledge. Morgan {\&} Claypool Publishers, 2021.
\newblock ISBN 978-3-031-00790-3.
\newblock \doi{10.2200/S01125ED1V01Y202109DSK022}.
\newblock URL \url{https://doi.org/10.2200/S01125ED1V01Y202109DSK022}.

\bibitem[Jarnac and Monnin(2022)]{jarnacM22}
L.~Jarnac and P.~Monnin.
\newblock Wikidata to bootstrap an enterprise knowledge graph: How to stay on topic?
\newblock In \emph{Proceedings of the 3rd Wikidata Workshop 2022 co-located with the 21st International Semantic Web Conference (ISWC2022), Virtual Event, Hanghzou, China, October 2022}, volume 3262 of \emph{{CEUR} Workshop Proceedings}. CEUR-WS.org, 2022.
\newblock URL \url{https://ceur-ws.org/Vol-3262/paper16.pdf}.

\bibitem[Jarnac et~al.(2023)Jarnac, Couceiro, and Monnin]{jarnacCM23}
L.~Jarnac, M.~Couceiro, and P.~Monnin.
\newblock Relevant entity selection: Knowledge graph bootstrapping via zero-shot analogical pruning.
\newblock In \emph{Proceedings of the 32nd {ACM} International Conference on Information and Knowledge Management, {CIKM} 2023, Birmingham, United Kingdom, October 21-25, 2023}, pages 934--944. {ACM}, 2023.
\newblock \doi{10.1145/3583780.3615030}.
\newblock URL \url{https://doi.org/10.1145/3583780.3615030}.

\bibitem[Lehmann et~al.(2015)Lehmann, Isele, Jakob, Jentzsch, Kontokostas, Mendes, Hellmann, Morsey, van Kleef, Auer, and Bizer]{lehmannIJJKMHMK15}
J.~Lehmann, R.~Isele, M.~Jakob, A.~Jentzsch, D.~Kontokostas, P.~N. Mendes, S.~Hellmann, M.~Morsey, P.~van Kleef, S.~Auer, and C.~Bizer.
\newblock Dbpedia - {A} large-scale, multilingual knowledge base extracted from wikipedia.
\newblock \emph{Semantic Web}, 6\penalty0 (2):\penalty0 167--195, 2015.
\newblock \doi{10.3233/SW-140134}.
\newblock URL \url{https://doi.org/10.3233/SW-140134}.

\bibitem[Lim et~al.(2019)Lim, Prade, and Richard]{lim19}
S.~Lim, H.~Prade, and G.~Richard.
\newblock Solving word analogies: {A} machine learning perspective.
\newblock In \emph{Symbolic and Quantitative Approaches to Reasoning with Uncertainty, 15th European Conference, {ECSQARU} 2019, Belgrade, Serbia, September 18-20, 2019, Proceedings}, volume 11726 of \emph{Lecture Notes in Computer Science}, pages 238--250. Springer, 2019.
\newblock \doi{10.1007/978-3-030-29765-7_20}.
\newblock URL \url{https://doi.org/10.1007/978-3-030-29765-7_20}.

\bibitem[Liu et~al.(2023)Liu, Chabot, Troncy, Huynh, Labb{\'{e}}, and Monnin]{liuCTHLM23}
J.~Liu, Y.~Chabot, R.~Troncy, V.~Huynh, T.~Labb{\'{e}}, and P.~Monnin.
\newblock From tabular data to knowledge graphs: {A} survey of semantic table interpretation tasks and methods.
\newblock \emph{Journal of Web Semantics}, 76:\penalty0 100761, 2023.
\newblock \doi{10.1016/J.WEBSEM.2022.100761}.
\newblock URL \url{https://doi.org/10.1016/j.websem.2022.100761}.

\bibitem[Mahdisoltani et~al.(2015)Mahdisoltani, Biega, and Suchanek]{mahdisoltaniBS15}
F.~Mahdisoltani, J.~Biega, and F.~M. Suchanek.
\newblock {YAGO3:} {A} knowledge base from multilingual wikipedias.
\newblock In \emph{Seventh Biennial Conference on Innovative Data Systems Research, {CIDR} 2015, Asilomar, CA, USA, January 4-7, 2015, Online Proceedings}. www.cidrdb.org, 2015.
\newblock URL \url{http://cidrdb.org/cidr2015/Papers/CIDR15_Paper1.pdf}.

\bibitem[Miclet et~al.(2008)Miclet, Bayoudh, and Delhay]{miclet2008}
L.~Miclet, S.~Bayoudh, and A.~Delhay.
\newblock Analogical dissimilarity: Definition, algorithms and two experiments in machine learning.
\newblock \emph{Journal of Artificial Intelligence Research}, 32:\penalty0 793--824, 2008.
\newblock \doi{10.1613/jair.2519}.
\newblock URL \url{https://doi.org/10.1613/jair.2519}.

\bibitem[Mitchell(2021)]{mitchell21}
M.~Mitchell.
\newblock Abstraction and analogy-making in artificial intelligence.
\newblock \emph{Annals of the New York Academy of Sciences}, 1505\penalty0 (1):\penalty0 79--101, 2021.

\bibitem[Noy et~al.(2019)Noy, Gao, Jain, Narayanan, Patterson, and Taylor]{noyGJNPT19}
N.~F. Noy, Y.~Gao, A.~Jain, A.~Narayanan, A.~Patterson, and J.~Taylor.
\newblock Industry-scale knowledge graphs: lessons and challenges.
\newblock \emph{Communications of the {ACM}}, 62\penalty0 (8):\penalty0 36--43, 2019.
\newblock \doi{10.1145/3331166}.
\newblock URL \url{https://doi.org/10.1145/3331166}.

\bibitem[Pan et~al.(2024)Pan, Luo, Wang, Chen, Wang, and Wu]{panLWCWW24}
S.~Pan, L.~Luo, Y.~Wang, C.~Chen, J.~Wang, and X.~Wu.
\newblock Unifying large language models and knowledge graphs: A roadmap.
\newblock \emph{IEEE Transactions on Knowledge and Data Engineering}, page 1–20, 2024.
\newblock ISSN 2326-3865.
\newblock \doi{10.1109/tkde.2024.3352100}.
\newblock URL \url{http://dx.doi.org/10.1109/TKDE.2024.3352100}.

\bibitem[Sequeda and Lassila(2021)]{sequedaL2021}
J.~Sequeda and O.~Lassila.
\newblock \emph{Designing and Building Enterprise Knowledge Graphs}.
\newblock Synthesis Lectures on Data, Semantics, and Knowledge. Morgan {\&} Claypool Publishers, 2021.
\newblock \doi{10.2200/S01105ED1V01Y202105DSK020}.

\bibitem[Shbita et~al.(2023)Shbita, Gentile, Li, DeLuca, and Ren]{shbitaGLDR23}
B.~Shbita, A.~L. Gentile, P.~Li, C.~DeLuca, and G.~Ren.
\newblock Understanding customer requirements - an enterprise knowledge graph approach.
\newblock In \emph{The Semantic Web - 20th International Conference, {ESWC} 2023, Hersonissos, Crete, Greece, May 28 - June 1, 2023, Proceedings}, volume 13870 of \emph{Lecture Notes in Computer Science}, pages 625--643. Springer, 2023.
\newblock \doi{10.1007/978-3-031-33455-9_37}.

\bibitem[Swartout et~al.(1996)Swartout, Patil, Knight, and Russ]{swartoutPKR1996}
B.~Swartout, R.~Patil, K.~Knight, and T.~Russ.
\newblock Toward distributed use of large-scale ontologies.
\newblock In \emph{Proceedings of the Tenth Workshop on Knowledge Acquisition for Knowledge-Based Systems}, volume 138, page~25, 1996.

\bibitem[Tanon et~al.(2020)Tanon, Weikum, and Suchanek]{tanonWS20}
T.~P. Tanon, G.~Weikum, and F.~M. Suchanek.
\newblock {YAGO} 4: {A} reason-able knowledge base.
\newblock In \emph{The Semantic Web - 17th International Conference, {ESWC} 2020, Heraklion, Crete, Greece, May 31-June 4, 2020, Proceedings}, volume 12123 of \emph{Lecture Notes in Computer Science}, pages 583--596. Springer, 2020.
\newblock \doi{10.1007/978-3-030-49461-2_34}.
\newblock URL \url{https://doi.org/10.1007/978-3-030-49461-2_34}.

\bibitem[Vrandecic and Kr{\"{o}}tzsch(2014)]{vrandecicK14}
D.~Vrandecic and M.~Kr{\"{o}}tzsch.
\newblock Wikidata: a free collaborative knowledgebase.
\newblock \emph{Communications of the {ACM}}, 57\penalty0 (10):\penalty0 78--85, 2014.
\newblock \doi{10.1145/2629489}.
\newblock URL \url{https://doi.org/10.1145/2629489}.

\bibitem[Weikum et~al.(2021)Weikum, Dong, Razniewski, and Suchanek]{weikumDRS21}
G.~Weikum, X.~L. Dong, S.~Razniewski, and F.~M. Suchanek.
\newblock Machine knowledge: Creation and curation of comprehensive knowledge bases.
\newblock \emph{Foundations and Trends Databases}, 10\penalty0 (2-4):\penalty0 108--490, 2021.

\bibitem[Zuckerman()]{zuckerman21}
L.~Zuckerman.
\newblock Tracking looted art with graphs: A case study.
\newblock URL \url{https://api.semanticscholar.org/CorpusID:247314664}.

\bibitem[Zuckerman(2020)]{zuckerman20}
L.~Zuckerman.
\newblock Linked data and holocaust era art markets: Gaps and dysfunctions in the knowledge supply chain.
\newblock In \emph{Proceedings of the International Conference Collect and Connect: Archives and Collections in a Digital Age, Leiden, the Netherlands, November 23-24, 2020}, volume 2810 of \emph{{CEUR} Workshop Proceedings}, pages 13--24. CEUR-WS.org, 2020.
\newblock URL \url{https://ceur-ws.org/Vol-2810/paper2.pdf}.

\end{thebibliography}

\end{document}